\documentclass[final,5p,times,twocolumn]{elsarticle} 

\usepackage{amssymb, amsmath, graphicx}
\usepackage[colorlinks=true, allcolors=blue]{hyperref}

\begin{document}

\begin{frontmatter}

\title{Boltzmann State-Dependent Rationality}

\author{Osher Lerner \\
University of California, Berkeley}
\begin{abstract}
    This paper expands on existing learned models of human behavior via a measured step in structured irrationality.
    Specifically, by replacing the suboptimality constant $\beta$ in a Boltzmann rationality model with a function over states $\beta(s)$, we gain natural expressivity in a computationally tractable manner.
    This paper discusses relevant mathematical theory, sets up several experimental designs, presents limited preliminary results, and proposes future investigations.
\end{abstract}

\end{frontmatter}


\section{Introduction}

With the eruption of deep learning we've created many strong "intelligence" engines.
Yet despite impressive results in academia, there exists a great gap in deploying systems in the world alongside humans.
Currently, operational frameworks for human-computer interaction are the missing link, and a key component of these frameworks are models of human behavior.

In a collaboration setting, simple algorithms often overestimate human rationality. The existence of a mathematically clear optimal sequence of actions does not guarantee that a spatially and temporally imprecise operator with a unique belief state will perform them. 
As collaborating humans, we temper our expectations, communication, and planning according to task difficulty, our partner's comfort and acuity with the task, and the uncertainty in their information and actions. 
In order to successfully collaborate with us, robots must --- at least implicitly --- model these notions.

A standard method for modeling our suboptimality is "Boltzmann Rationality" \cite{MaxEnt}, which models a noisy optimal trajectory based on exponentially weighting the cost of available trajectories. However, this model still falls short in accurately modeling human states.
Research on models of systematic suboptimality, such as the "Boltzmann Policy Distribution" \cite{BPD}, if successful, could open the flood-gates towards countless downstream competitive, assistive, collaborative, and observational methods.
Furthermore, we believe a better understanding of human suboptimality could provide insight into how we represent and plan tasks.
In that manner, if we learn how humans are stupid, along the way we may learn useful frameworks for how we are so smart.



\section{Background}

Under the moniker "Industry 4.0", the next industrial revolution is expected to be the collaboration of humans and robots \cite{Cognitive}.
When we combine the consistency, precision, and power of robots with the dexterity, intuition, and general adaptive knowledge of humans, our industries are able to operate far more effectively than using only one of the two.
A hot area of research is thus the efficient coordination of fleets of humans and robots. 
Much work has been published on the management of dynamic new safety conditions \cite{Safety}, fatigue \cite{Cognitive}, ergonomics, skill \cite{TaskAlloc}, supervision, cycle times, preference \cite{TaskChar}, between humans, robots, and teams of humans and/or robots in assembly tasks.
However, current work is often rigidly fixed to hand-specified characteristics and thus unable to adapt to general considerations like humans.
The space of strategies is vast, especially since collaboration can grow to great industrious scales. 
\cite{Proactive} outlines the myriad space of considerations and methods and concludes that in the 2020s we will soon reach levels beyond even support, coordination, and cognition on to co-evolution.
In this imagined system, robots initiate proactive actions, agents are self organizing, and cognition and empathy are shared amongst both humans and robots.
This structure requires accounting for numerous models of mutual prediction, perception, uncertainty estimation, and well-being.

One key such model is human proficiency or task difficulty. This notion appears to critically affect other aspects of collaboration tasks, such as the importance of nonverbal communication \cite{Nonverbal} and the asymmetry of updates to trust \cite{MultiTaskPaper}.

\subsection{Models of Human Proficiency}

In many supervised learning settings, machine learning models have reached human ability, and now seek to go beyond. For this purpose, researchers have modeled the suboptimality of the human labelers. In 
\cite{Crowds}, they learn each human annotator's decision parameters inside a (simultaneously learned) latent image representation space.
This simultaneous optimization successfully reveals human competence, difficulty of classification, and superhuman annotations --- all three of which are image-dependent.

In the context of RL, models of capability have been developed for robots, since human operators often misunderstand joint limits of robotic arms.
\cite{ShowMe} introduces a framework including a function $b_h$ mapping a state to a value between 0 and 1 representing a human's belief of how likely it is to be reachable by the robot.
This formalism is then flipped around to optimize demonstrations to best update the human's beliefs, and achieves better performance than simply sampling from untraversed waypoints.

In \cite{ILEED}, the authors use \cite{ObsMLE}'s annotator error maximum likelihood estimation algorithm for imitation learning, surpassing rewards achieved by SOTA IL algorithms in GridWorld, robotic manipulation, and chess endgame settings.
They embed demonstrators and states in a shared representation space and calculate "expertise level" as a similarity of the two, and theoretically show they can recover the true optimal policy from suboptimal demonstrations.

\subsection{Inverse RL}
Given an environment with states $S$ (often partial observations), actions $A$, and rewards $R$ for state transitions, the reinforcement learning problem is to find a policy $\pi(a \mid s)$ optimizing expected returns of a trajectory $\xi$ sampled from the environment.
\begin{align*}
    \pi^* = \arg\max_\pi \mathbb{E}_{\xi \sim \pi}[U(\xi)]
\end{align*}
IRL solves the inverse problem. Given an environment and rollout data of trajectories of a certain policy, we want to calculate the cost function $U(\xi)$ that the policy is maximizing. 
The space of cost functions is vast, so this problem is under-determined given a finite number of samples. We wish to find meaningful solutions, which depending on the context can mean interpretable feature weights, well-conditioned for optimization, and/or similarity to the true cost function (under some notion of similarity in cost function space).

In practice, we observe data from policies other than the optimal policy $\pi^*$. Most often for HRI tasks, the data we collect is of humans who are optimizing an objective subject to context clues, internal beliefs, etc. Relative to robot action spaces, human actions have imprecision in space and time, and are roughly planned rather than exactly optimal. Algorithms solving the IRL problem for HRI tasks must account for human suboptimality in its many forms.

\subsection{Boltzmann Rationality}

The Boltzmann Rationality (BR) model over trajectories is formulated as the solution to the near-optimal maximum entropy IRL problem.

That is,
\begin{align*}
    \max_P H(P) \,\,\,\,\,\,\,\,\,\,\text{ s.t. }\, \mathbb{E}_{\xi_D \sim P} [U(\xi_D)] \approx \min_\xi U(\xi) 
\end{align*}
is solved by the following Boltzmann Rational distribution over trajectories:

\begin{align}
    P(\xi) &= \frac{1}{Z} e^{-U(\xi)} \label{Boltz-simple}
\end{align}
where $Z$ is a constant normalization factor, and $U(\xi)$ is the cost accumulated by the trajectory $\xi$.
Typically, cost is calculated as a function of each state, and summed over the trajectory. Although a discount factor is sometimes used, for simplicity we will use $\gamma = 1$.
The parametrization of the cost function is taken to be a vector of weights weighing each feature of the state, computed by some map $\phi(s)$. These features can be prescribed, learned offline, or updated online.
\begin{align}
    P(\xi \mid \theta) &= \frac{1}{Z} e^{-\sum_{s \in \xi} \theta^T \phi(s)}
\end{align}

In order to account for a range of possible "distances" from optimality, an "inverse temperature" parameter $\beta$ is introduced. Varying $\beta$ will interpolate between the optimal policy $\beta = \infty$ and uniform policy $\beta = 0$
(though differently from action-space interpolation such as used in  \cite{ILEED}).

\begin{align}
    P(\xi \mid \theta, \beta) &= \frac{1}{Z} e^{-\beta \sum_{s \in \xi} \theta^T \phi(s)} \label{Boltz-rat}
\end{align}
Note that any distribution can be induced from Equation \ref{Boltz-simple} by construing a particular cost function – which is precisely the problem posed by IRL – and so one might assume any new parameters would be redundant.
However, by imposing structure on our cost and introducing new parameters, we aim to reach a more natural parametrization. Such a description can critically ease the optimization search and can encode interpretable information about what is being learned.
This core idea is integral to the algorithm presented in the next section.

The Boltzmann Rationality model seems very natural, as it coincides with generic theoretical derivations and many physical systems.
In practice however, the mismatch between this model and actual human behavior bottlenecks human-robot interaction algorithms, particularly in the case of collaboration.
The problem to which this paper contributes is finding better models of human behavior.

\section{Theory}

In this section, we introduce mathematical formulation of state dependent rationality. The derivations begin at the most general form, and we state any simplifying assumptions along the way.

To better model human behavior, we will introduce new parameters to describe systematic suboptimality, expanding on the one scalar value $\beta$.
First, we will impose additional structure on our cost, just as we did under Boltzmann Rationality. 
We will now consider several human agents, assuming they are all optimizing the same cost function. In an experiment with a clearly communicated objective where humans are subjected to the same task, this is a fair assumption.
Then, we will introduce the possibility that each human has varying suboptimality over states.

\subsection{Why States?}

It is not obvious that suboptimality should be formulated as a function over states. Often a notion of proficiency of a "maneuver" is preferred, while in other cases, a dynamic notion of attention, knowledge, and mood state of a human is needed.
The space of trajectories can be more expressive than states, but computations over it are usually intractable.
In different instantiations of the reinforcement learning framework, the relevant and rich information could be in the action space. In practice this is rarely the case, but a meaningful space to work in is the policy space, as is used for systematic suboptimality in BPD \cite{BPD}.

While these methods should all be investigated, formulating suboptimality over states however seems initially the most promising.
In many cases states themselves are the best available representation of when the environment may become noisy or unfamiliar. They also contain implicit information about trajectories and maneuvers, as certain states are only reached in specific traversals towards goals. Particularly, we can make use of the iterative nature of our optimizers to incorporate the implicit sequential nature of state data into our training algorithms. Notably, the alignment of state-based formulation with our data structure, cost formulation, and iterative training method means we can leverage many of the same conventional computational tricks to make optimization tractable.

\subsection{Forward Model}



Let's see how our trajectory distribution looks when we vary $\beta$ over states $s$.
We will parametrize this function just as we did the cost, as a vector of feature weights $\theta_\beta$. It may be useful to set or learn a separate featurization than the one used for rewards, but we will assume the features are descriptive enough to compute both suboptimality and rewards.
Equation \ref{Boltz-rat} becomes



\begin{align*}
    P(\xi \mid \theta,\beta) &= \frac{1}{Z} e^{\sum_{s \in \xi}\beta(s) \theta^T \phi(s)}  \nonumber \\
    P(\xi \mid \theta_R,\theta_\beta) &= \frac{1}{Z} e^{-\sum_{s \in \xi}\theta_\beta^T \phi_\beta(s) \theta_R^T \phi_R(s)}  \nonumber \\
    P(\xi \mid \theta_R,\theta_\beta) &= \frac{1}{Z} e^{-\sum_{s \in \xi}\theta_\beta^T \phi(s) \theta_R^T \phi(s)}  \nonumber \\
    P(\xi \mid \theta_R,\theta_\beta) &= \frac{1}{Z} e^{-\sum_{s \in \xi}\theta_\beta^T \phi(s) \phi(s)^T \theta_R }  \nonumber \\
    P(\xi \mid \theta_R,\theta_\beta) &= \frac{1}{Z} e^{-\theta_\beta^T \left(\sum_{s \in \xi} \phi(s) \phi(s)^T\right) \theta_R }  
\end{align*}

where 
\begin{equation}
    Z = \sum_{\bar{\xi}\in \Xi} e^{-\theta_\beta^T \left(\sum_{s \in \bar{\xi}} \phi(s) \phi(s)^T\right) \theta_R }
\end{equation}

To simplify notation, we define $\Phi_\xi = \sum_{s \in \xi} \phi(s) \phi(s)^T$ to be the "feature counts" matrix of trajectory $\xi$.
So our trajectory distribution is 
\begin{equation} \label{traj-dist}
    P(\xi \mid \theta_R,\theta_\beta) = \frac{1}{Z} e^{-\theta_\beta^T \Phi_\xi \theta_R } 
\end{equation}

\subsection{Inverse Model}
Now we invert the reinforcement learning problem, which can be computed using Bayesian inference.
Let's consider several humans with the same reward model but varying state-dependent proficiency.
Given rollouts $\xi^i_j$ from human $i$ and run $j$, we compute
\begin{align}
    P(\theta_R,\{\theta^i_\beta\} \mid \{\Xi^i\}) &= \frac{P(\{\Xi^i\} \mid \theta_R,\{\theta^i_\beta\}) P(\theta_R,\{\theta^i_\beta\})}{P(\{\Xi^i\})} 
\end{align}
We assume the rollouts are independent of each other up to our human parameters.
Note, this assumption could be broken if the humans gain proficiency during data collection (between or during trials).
Mathematically, this means 
\begin{align*}
    P(\{\Xi^i\} \mid \theta_R,\{\theta^i_\beta\}) = \prod_i P(\Xi^i \mid \theta_R,\theta^i_\beta) = \prod_i \prod_j P(\xi^i_j \mid \theta_R,\theta^i_\beta)
\end{align*}
We also assume the human's parameters are independent of each other, except for reward which they share.
\begin{align*}
    P(\theta_R,\{\theta^i_\beta\}) &= P(\theta_R)\prod_i P(\theta^i_\beta)
\end{align*}
By plugging these derived equations in, we will now attempt to solve for the parameters $\theta^*$ (where $\theta$ refers to $\theta_R$ and $\{\theta_\beta^i\}$) that achieve the maximum likelihood.


\begin{align}
    \theta^*
    &= \arg\max_\theta\,\, P(\theta \mid \{\Xi^i\}) \nonumber \\
    &= \arg\max_\theta\,\, \log P(\theta \mid \{\Xi^i\}) \nonumber \\
    &= \arg\max_\theta\,\, \log P(\{\Xi^i\} \mid \theta) + \log P(\theta) - \log P(\{\Xi^i\}) \nonumber \\
    &= \arg\max_\theta\,\, \log P(\{\Xi^i\} \mid \theta) + \log P(\theta) \nonumber \\
    &= \arg\max_\theta\,\, \sum_i \sum_j \log P(\xi^i_j \mid \theta_R,\theta^i_\beta) + \log P(\theta) \nonumber \\
    &= \arg\min_\theta\,\, \sum_i \sum_j \left(\theta_R^T \Phi_{\xi^i_j} \theta_\beta^i + \log Z(\theta)\right)-  \log P(\theta) \label{MLE}
\end{align}

This optimization problem is theoretically analyzed in \ref{app-argmax}.

\section{Experiments}

To test the applicability of our theory, there are several experiments we can try.
Unfortunately we failed to run these experiments to produce results, but the rough experimental design for these is elaborated here.

\subsection{Environments}
We run a simple GridWorld environment to tractably test our math. 
For the OverCooked setting, we use the environment provided in \cite{Overcooked} and human data gathered from Mechanical Turk \cite{OvercookedData} labeled with human IDs.
We chunk our data by layout and human ID, with a total of 8 different tasks, 88 humans, and 8 (sometimes less) rollouts from each agent of 397 timesteps each.
Note this data is all gathered in the collaborative setting, but we are learning only one agent's reward and suboptimality parameters, and absorbing the other into the environment.

\subsection{Parameter recovery}
\begin{figure}[h!]
	\centering    
    \fbox{\includegraphics[width=0.2\textwidth]{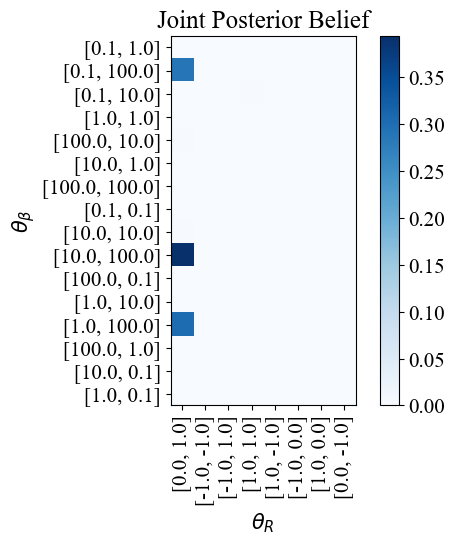}}   
    \hspace{30px}
    \fbox{\includegraphics[width=0.2\textwidth]{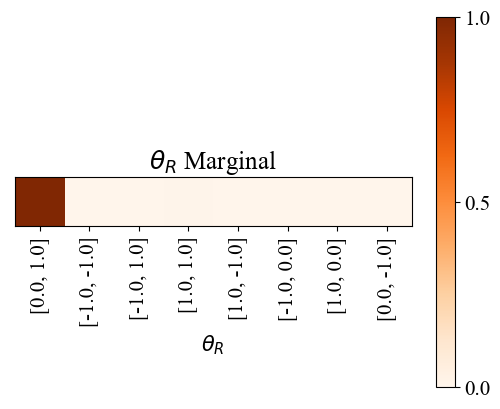}}
    \hspace{3px}
    \fbox{\includegraphics[width=0.2\textwidth]{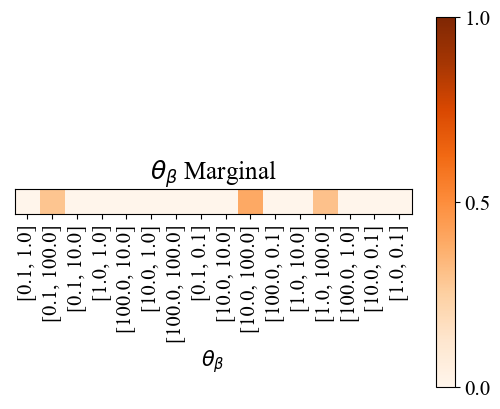}}
	\caption{Posterior Belief of $\theta$ parameters from trajectories generated by $\theta_R = [0, 1]$ and $\theta_\beta = [1, 100]$. 
    For this arbitrary discrete space, we are able to accurately recover our parameters.} 
    \label{posteriors}
\end{figure}

First, we want to ensure our math is internally consistent, and that our parameter spaces are meaningful.
For this purpose, we create agents operating directly using our model, and see if we can recover their internal parameters just from rollout observation.
In the GridWorld setting, this seems works pretty well (see Figure \ref{posteriors}), though a compilation of rigorous results has not been collected. 

\subsection{Shared Goals}
We can construct another experiment to test our assumption of shared goals.
We assumed that by communicating to each play-tester the same objective, their actions would align with models that share state feature reward weights. 
For any task in which we ground our training from trials on the same objective, we can validate this assumption experimentally.
Using live subjects, we can ask the humans to label sampled trajectories from different human runs and ask them to choose the trajectory that best achieves the goal of the task, and see if there exists a statistical difference between the human labelers.

With offline human data (such as our MTurk dataset), we can at least qualitatively (if not statistically) compare the learned reward parameters of trajectories by different humans learned with BR to see how closely they align compared to those learned from other tasks and random weights.

\subsection{Generalization of Learned Rewards}
We wish to test how well our learned $\theta_R$ generalizes, to show it meaningfully corresponds to the reward model and is independent from human suboptimality.


We set up an experiment on each of our representative tasks (from Overcooked and GridWorld). 
We fit a BR rational model to the data in aggregate and a BSDR model with each human identity labeled.
Then, we optimize the learned $\theta_R$ to find $\pi^*_{\theta_R}$. 
By measure the performance of these learned policies with the original task rewards (known to the experimenters, but not our algorithms), we can see which achieves better performance. The cost function that can be opimized to achieve a higher true reward should be more closely aligned to the true reward.

We can also compare the rewards predicted by our learned parameters evaluated over some sample human trajectories to the true values.

Note that it seems BSDR has a competitive advantage in this experiment since it gets an extra dimension of data: human ID's.
Without imposing the structure of different agents with shared cost, our BSDR parameters are redundant and have no preference for one corresponding to reward and the other to suboptimality. Thus BSDR cannot generalize from training on only 1 human.
To equalize the playing field, we can train BR with different $\beta_i$ for each human, and evaluate their performance individually.

\subsubsection{Action Prediction}

In this experiment, we directly compare the action distributions predicted by our human models.
Under our list of tasks, we again train BR, BPD, and BSDR.
We then compare their cross entropy prediction performance of human data to each other.
We can also plot results for a self-play policy, and a random policy.

\subsection{Generalization of Learned Suboptimality for Goal Inference}


After learning from demonstrator's performing the same tasks, we should be able to use our human suboptimality models to better infer new goals and predict their actions.



\subsection{Goal Inference}
In this experiment, we test the ability to predict a goal from a partial trajectory, which should be easier if we have a pre-existing accurate model of their suboptimality.

We use our calculated values of 
$\beta^i$ and $\theta_\beta^i$ from the last experiment.
We restrict the reward parameters to be one of a small set. In this case, we will use GridWorld with certain coordinates as goals.
Then, we consider unseen trajectories by the same human used for training the $\beta$ parameters.
Using only the initial portion of the trajectory, we numerically compute with Bayesian inference the likelihood of each possible goal.
We plot the likelihood of the true goal computed from BR and BSDR averaged over 8 different trajectories from 20 different humans. We repeat the trial for the first 25\%, 50\%, 75\%, and 100\% of the trajectory. And we again repeat for different GridWorld environments.

Unfortunately, we did not yet collect human data on GridWorld tasks.



\section{Future Steps}

\begin{itemize}
\item State-dependent suboptimality may be useful in certain in environments with relevant information encoded in the state, while irrelevant in others.
Learning a latent representation of states, actions, or maneuvers may be the best bet. 
\cite{ILEED} cited meaningful state features as challenging to learn for unexplored environments, but 
\cite{TRAIL} successfully learned abstract state-dependent action representations to surpass demonstrator performance. 
Once we have a meaningful rich space we can learn suboptimality as a function of that representation.

\item One could also investigate the use of more complex suboptimality models such as different ways of interpolating between optimality and irrationality.

\item To best gauge human models, we need to work with real human data. Diverse tasks, environments, and agents are important for experiments aiming to understand human models.

\item The question is left of how to take advantage of these human models for collaboration.
Knowledge of $\beta(s)$ could be used to estimate conditional uncertainty to assess risk in plans.
For assistance tasks where the human's objective is unknown, an assistance task could be to steer the state towards the human's expertise.
In a game context, given the human model, one could explicitly compute the best response strategy to it.

\item We can experiment using different features for our reward and suboptimality models.
\end{itemize}


\section*{Acknowledgements}
Anca Dragan and Cassidy Lailaw, the instructors of CS 287H at Berkeley, both authored background papers, brainstormed on this idea, gave feedback, and supported me in writing this paper. For them I am very grateful.

\appendix

\section{Further Analysis of the Max Likelihood Problem} \label{app-argmax}

The following is an incomplete attempt at analytically analyzing the optimization problem from the MLE in Equation \ref{MLE}.
It can be safely ignored.

Note that the $Z(\theta)$ term is omitted from the following computations, which is an error.

\begin{align*}
    \theta^*
    &= \arg\max_\theta \sum_i \sum_j \left(\theta_\beta^T \left(\sum_{s \in \xi^i_j} \phi(s) \phi(s)^T\right) \theta_R\right) + \log P(\theta_R) + \sum_i \log P(\theta^i_\beta)
\end{align*}

This quantity can be computed efficiently by flattening $\Xi^i$ into a vector of states visited, and performing tensor multiplications with the $\theta^i$ vector. An einsum summation could work as well.
Notice that the terms involving $\theta^i$ can be separated out. So 
\begin{align*}
    \theta^* 
    &= \arg\min_{\theta^i_\beta} \left(\sum_{s \in \Xi^i} \theta_R^T \phi(s)\phi(s)^T\right) \theta_\beta^i  + \log P(\theta^i_\beta) 
\end{align*}
We can summarize our data with the frequency rates at which human $i$ visits state $s$: $\rho^i(s)$.
Then   
Given a discretized search space of $\theta$ values we can simply search over them all.

To find this numerically for large spaces, let's compute the gradient.

Let's define the matrix $\Phi^i = \sum_j \sum_{s \in \xi^i_j} \phi(s) \phi(s)^T$.

Let's assume a prior that is uniform over $\theta$ with unit length, and 0 otherwise. 
Then our maximum likelihood problem becomes
\begin{align*}
    \theta^* 
    &= \arg\min_{\|\theta\| = 1} \theta_R^T\sum_i \Phi^i \theta_\beta^i 
\end{align*}

Then we see that $\theta_R$ that minimizes the quantity is just the unit vector in the opposite direction from 
$\sum_i \Phi^i \theta_\beta^i$.
$$\theta_R^* = -\frac{\sum_i \Phi^i \theta_\beta^i}{\|\sum_i \Phi^i \theta_\beta^i\|}$$
Then with that fixed, we can go into the minimization over $\{\theta_\beta^i\}$.
We are left with 
\begin{align*}
    \theta^* &= \arg\max_\theta P(\theta_R,\{\theta^i_\beta\} \mid \{\Xi^i\}) \\
    &= \arg\min_{\|\theta\| = 1} -\frac{1}{\|\sum_i \Phi^i \theta_\beta^i\|}(\sum_i \Phi^i \theta_\beta^i)^T(\sum_i \Phi^i \theta_\beta^i) \\
    &= \arg\min_{\|\theta\| = 1} -\frac{1}{\|\sum_i \Phi^i \theta_\beta^i\|}\|\sum_i \Phi^i \theta_\beta^i\|^2 \\
    &= \arg\max_{\|\theta\| = 1} \|\sum_i \Phi^i \theta_\beta^i\| \\
    &= \arg\max_{\|\theta\| = 1} \|\sum_i \Phi^i \theta_\beta^i\|^2
\end{align*}

This is not an obvious expression to maximize, but we can get some insight to the theoretical analysis. Let's calculate the derivative of the unconstrained objective
\begin{align*}
    \frac{\partial}{\partial \theta_\beta^i} \|\sum_i \Phi^i \theta_\beta^i\|^2
    &= \frac{\partial}{\partial \theta_\beta^i} (\sum_i \Phi^i \theta_\beta^i)^T (\sum_{i'} \Phi^{i'} \theta_\beta^{i'}) \\
    &= 2(\frac{\partial}{\partial \theta_\beta^i} \sum_i \Phi^i \theta_\beta^i)^T (\sum_{i'} \Phi^{i'} \theta_\beta^{i'}) \\
    &= 2{\Phi^i}^T (\sum_{i'} \Phi^{i'} \theta_\beta^{i'}) \\
\end{align*}

The critical points appear when the derivatives are all 0. This occurs at a minimum when $\sum_i \Phi^i \theta_\beta^i = 0$.
Other critical points appear whenever $\sum_i \Phi^i \theta_\beta^i$ is in the null space of every ${\Phi^i}^T$. Given our weight constraint,  
we can use Lagrange multipliers to solve the constrained optimization problem.
Let $f(\theta) = \|\sum_i \Phi^i \theta_\beta^i\|^2$ and $g(\theta) = \|\theta_\beta^i\| - 1$.
Then our likelihood is maximized when $L(\theta, \lambda) = f(\theta) + \lambda g(\theta)$ is at a stationary point.

\begin{align*}
    \forall_i \,\,\,\,\,\frac{\partial}{\partial \theta_\beta^i} L(\theta, \lambda) = 0 \,\,\,&\,\,\,\,\,\,\,\,\,\,\,\,\,\,\,\,\,
    \frac{\partial}{\partial \lambda} L(\theta, \lambda) = 0 \\
    \forall_i \,\,\,\,\,0 &= 2{\Phi^i}^T (\sum_{i'} \Phi^{i'} \theta_\beta^{i'}) + 2 \lambda \theta_\beta^{i} \ \\
    1 &= \|\theta_\beta^i\| \\
\end{align*}

These computations can in practice be rather large.
\begin{itemize}
    \item With a large state dimension $D$, $\Phi^i$ will have $D^2$ entries. It may be easier to keep the expression decomposed into $\sum_{s \in \Xi^i} \theta_R^T \phi(s)\phi(s)^T$.
    \item For limited data per human, it may be more efficient to use the sum over states.
    \item In the case of large data per human, and a small state space (or in rare cases, a small number of states visited per human), it can be efficient to transform the sum over states into a product of a state feature tensor by $\rho^i(s)$, a vector representing the frequency at which a human visits state $s$. We can construct an operational quantity like that used in MaxEntIRL: feature counts $\tilde{\phi}^i$.
\end{itemize}

\bibliographystyle{alpha}
\bibliography{bibliography}

\end{document}